\documentclass{article} 
 \newlength\titlebox 
\usepackage{times}  
\usepackage{helvet} 
\usepackage{courier}  
\usepackage[hyphens]{url}  
\usepackage{graphicx} 
\usepackage{natbib}  
\usepackage[switch]{lineno}
\usepackage[utf8]{inputenc} 
\usepackage[T1]{fontenc}    
\usepackage{url}            
\usepackage{booktabs}       
\usepackage{amsfonts}       
\usepackage{nicefrac}       
\usepackage{microtype}      
\usepackage{caption}
\usepackage{subcaption}
\usepackage[affil-it]{authblk}
\title{Risk Assessment for Machine Learning Models}
\author[1,*]{Paul Schwerdtner}
\author[1]{Florens Greßner}
\author[2]{Nikhil Kapoor}
\author[1]{\\ Felix Assion}
\author[2]{Ren\'e Sass}
\author[1]{Wiebke Günther}
\author[2]{Fabian Hüger}
\author[2]{Peter Schlicht}

\affil[1]{Neurocat GmbH}
\affil[2]{Volkswagen AG}
\affil[*]{Corresponding author: \texttt{ps@neurocat.ai}}

\usepackage{amsmath}
\usepackage{bbm}
\usepackage{amsfonts}
\usepackage{pgfplots,tikz}
\usetikzlibrary{external,arrows.meta,calc,shapes,pgfplots.groupplots,decorations.markings,graphs,spy,intersections}
\newtheorem{defi}{Definition}
\newcommand{\R}{\mathbb{R}}
\newcommand{\dd}{\mathsf{d}}

\newcommand{\trc}{\bar{f}_c}
\newcommand{\nx}{n_{\mathcal{X}}}
\newcommand{\nd}{n_{\mathcal{D}}}

\tikzset{
    nonterminal/.style={
      rectangle,
      top color=black!10, bottom color=black!10, font = \bfseries,minimum size=10mm,draw=UniBlue,rounded corners},
    line/.style={-latex’},
    fom/.style={ line width=0pt,
      rectangle,minimum size=10mm,
      draw=black!10, top color=black!10,bottom color=black!10},
    rom/.style={ line width=0pt,
      rectangle,minimum size=9mm,
      draw=black!10, top color=black!10,bottom color=black!10},
      alg/.style={ line width=0pt, rectangle, draw=white, fill=black!10},
    sf/.style={rectangle, fill=black!10, minimum width=0.8cm, minimum height=0.2cm, outer sep=0.01cm, inner sep=0mm},
    tf/.style={rectangle, fill=black!10, minimum width=0.8cm, minimum height=0.8cm, outer sep=0.01cm, inner sep=0mm},
    ts/.style={rectangle, fill=black!10, minimum width=0.2cm, minimum height=0.8cm, outer sep=0.01cm, inner sep=0mm},
    ss/.style={rectangle, fill=black!10, minimum width=0.2cm, minimum height=0.2cm, outer sep=0.01cm, inner sep=0mm},
    min/.style={rectangle, fill=white, draw=white, text=black, minimum width=0.2cm, outer sep=0.01cm, inner sep=0mm},
    ssk/.style={rectangle, fill=UniBlue, minimum width=0.2cm, minimum height=0.2cm, outer sep=0.01cm, inner sep=0mm},
    action/.style={rectangle, fill=black!10, text width=3cm, align=center},
    decision/.style={rectangle, text width=3cm, fill=black!10!red!10, align=center, inner sep=1mm}
}

\begin{document}

\twocolumn
\maketitle
\begin{tikzpicture}[remember picture, overlay]
  \node[draw=white,thick,minimum width=\textwidth,text width=\textwidth] at ([yshift=-0.5in]current page.north) {\hfill Accecpted for presentation at the NeurIPS 2020 Virtual Workshop: Machine Learning for Autonomous Driving};
\end{tikzpicture}

\begin{abstract}
  In this paper we propose a framework for assessing the risk associated with deploying a machine learning model in a specified environment. For that we carry over the risk definition from decision theory to machine learning. We develop and implement a method that allows to define deployment scenarios, test the machine learning model under the conditions specified in each scenario, and estimate the damage associated with the output of the machine learning model under test. Using the likelihood of each scenario together with the estimated damage we define \emph{key risk indicators} of a machine learning model.

  The definition of scenarios and weighting by their likelihood allows for standardized risk assessment in machine learning throughout multiple domains of application. In particular, in our framework, the robustness of a machine learning model to random input corruptions, distributional shifts caused by a changing environment, and adversarial perturbations can be assessed.

\end{abstract}

\section{Introduction}

With the deployment of machine learning (ML) models in safety and security critical environments, risk assessment becomes a pressing issue. Failure modes of a given ML model must be identified and the likelihood of occurrence and severity of the damage caused by failure must be assessed. In this work, we focus on failures that result from input perturbations and provide a framework that allows to integrate different sources of input perturbations and to compute general risk scores for a given network and operational environment. These \emph{key risk indicators} (KRIs) can guide the decision on whether it is safe and secure to deploy a given ML model in a specified environment.

For the evaluation of ML risk, we consider \emph{adversarial} input data and \emph{corrupted} input data, which can be used to evaluate ML security and ML safety, respectively. In particular, to qualify as adversarial input data, we assume that a perturbation on the input is specifically crafted to maximize the difference between a ML model's output and the human interpretation of that same input. On the other hand, \emph{corrupted} input data is usually generated ML model agnostic and follows a somewhat \emph{natural} distribution of input data or naturally occurring noise. 

In recent years, it has become a well-known fact that neural networks (NN), a subset of ML models, are susceptible to adversarial perturbations \cite{goodfellow_explaining_2015} and various algorithms have been proposed to compute such perturbations effectively (known as \emph{adversarial attacks}). It is important to note that due to the transferability of attacks between NN that perform a similar task, an attacker does not need to have access to the attacked NN to successfully craft adversarial perturbations \cite{liu_delving_2017}. Furthermore, adversarial attacks are not merely a fragile phenomenon but can also be planted in the real world to fool NNs \cite{kurakin_adversarial_2017}.

Alongside adversarial attacks a large number of adversarial defenses that are designed to detect and/or mitigate the effect of adversarial noise have been proposed. However, typically a few months after a defense has been published, an attack that circumvents the detection and mitigation mechanism of that defense is found \cite{carlini_adversarial_2017,athalye_obfuscated_2018}.

This \emph{attack and defense arms race} has led to the introduction of formal verification algorithms of NNs such as the seminal work of \cite{katz_reluplex_2017}. These algorithms are used to verify that around a given set of input points the NN's output does not change for perturbations up to a certain size usually measured in either the $\ell_0$, $\ell_2$, or $\ell_\infty$ norm. However, such formal verification methods do not scale to larger, industry relevant tasks  without sacrificing rigor. Furthermore, realistic attack scenarios or image corruptions which are usually not bounded in some $\ell_p$ norm render the formal verification techniques inappropriate in these situations.

\begin{figure}[tb]
  \centering
  \begin{tikzpicture}[line width=1pt, >=Latex, inner sep=2mm, outer sep=0mm,every node/.style={scale=1.15}, node distance=2.5cm]
    \node (mlmod) [fom,minimum width=2.2cm] {\small{ML model}};
    \node (env) [fom, right of=mlmod,align=center,minimum width=2.2cm] {\small{Deployment}\\ \small{Scenarios}};
    \node (sev) [fom, right of=env,align=center,minimum width=2.2cm] {\small{Severity}\\ \small{Estimation}};
    \node (cube) [fom, below of=env, above=0.3cm, minimum width=7.2cm] {\small{Risk Tensor Computation}};
    \node (agr) [fom, below of=cube, above=0.3cm, minimum width=7.2cm] {\small{Aggregation and Filtering}};
    \node (kri) [fom, below of=agr, above=0.3cm, minimum width=7.2cm] {\small{Key Risk Indicators}};
    \draw [->] (mlmod.210) to node[auto] {$f_c$} (mlmod.210|-cube.north);
    \draw [->] (env.210) to node[auto] {$\mathcal{X}_j, \mathcal{D}_{x}$} (env.210|-cube.north);
    \draw [->] (sev.210) to node[auto] {$L_i$} (sev.210|-cube.north);
    \draw [->] (cube.south) to node[auto] {$R$} (cube.south|-agr.north);
    \draw [->] (agr.south) to node[auto] {$\hat \rho_i$} (agr.south|-kri.north);
  \end{tikzpicture}
  \caption{Overview of the proposed key robustness indicator computation method, which is explained in Section~\ref{sec:kritensor}}%
  \label{fig:overview}
\end{figure}
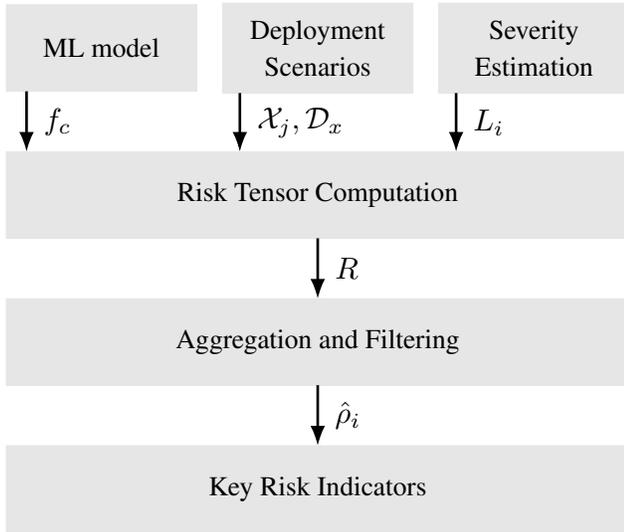

In \cite{tian_deeptest_2018} and \cite{pei_towards_2017} application-oriented robustness evaluation procedures were proposed that explicitly take a more realistic attack and corruption scope into account. As an example, instead of simply limiting the $\ell_2$ norm of a possible perturbation, the adversarial image transformation must be a rotation or change in brightness of the original image. The consideration of realistic image corruptions is key for risk assessment since a highly damaging perturbation that cannot occur in practice or, if at all, with vanishing probability, demands less action than a less problematic but still harmful perturbation that occurs regularly.

Therefore, we propose a framework that lets deployers of ML models define the possible perturbations and their respective magnitude and likelihood to set up realistic test scenarios. Then this scenario dependent robust performance is systematically evaluated by the introduction of KRIs. These indicators allow for comparability of ML models with respect to their robustness in different operational environments. This approach enables well-founded decisions on whether a ML model is fit for application. An overview of our method is given in Figure~\ref{fig:overview}. The input data consists of a ML model that is to be tested, previously designed deployment scenarios, and a severity estimation in the form of a loss function, that computes a damage associated with the ML model's output. From these inputs, we compute a risk tensor that is used as data-storage to be able to extract the required risk indicators by aggregation and filtering.

Next to estimating ML risk, our method can also be used to understand the failure modes of a ML model, or in particular the reason for success of implemented adversarial defenses. Specifically, throughout this paper, we make the following contributions.
\begin{itemize}
  \item We provide a framework in which the risk associated with deploying ML models in specified environments can be assessed in a standardized way.
  \item We provide a data-efficient tensor based method for storing robustness information on a given NN that can be queried and filtered to extract KRIs.
  \item We implement and test our framework on a set of image classifiers for the CIFAR10 dataset \cite{krizhevsky2009learning} to identify robustifying features of the training process or NN topology.
\end{itemize}

Our paper is organized as follows. In the next section we describe our setting and compare risk and robustness definitions common in ML to the risk definition from statistical decision theory. After that, we explain how we can apply the latter in the context of ML. For that we present a light-weight data structure that allows for scenario-based risk assessment of a ML model. We illustrate our method in an image classification case study, in which we identify the safest model for classifying images under a set of sensor, weather induced, random, and adversarial perturbations. 

\section{Background}
\label{sec:background}

We restrict our presentation of the theoretical background to classification as this allows for a more concise notation. However, it is important to note that our considerations immediately carry over to more complex tasks such as semantic segmentation or object detection.

We denote a classifier as $f_c : \R^{n_x}  \rightarrow \R^{n_c}$, where $n_x$ and $n_c$ are the length of the input vector (e.g.\ a vectorized input image) and the number of classes, respectively. Let $\mathcal{X}=(X,\mathcal{F},\mathbb{P})$ be the probability space of inputs and $\trc: \R^{n_x}  \rightarrow \R^{n_c}$ be the \emph{true} classifier that maps each input $x \in X$ to the correct class (in one-hot encoding).

The most common concept of (adversarial) ML robustness is based on the smallest perturbation necessary to provoke an incorrect classification \cite{fawzi_analysis_2018}, i.e.\
\begin{align*}
  \rho_1(f_c,\mathcal{X},\trc) := \mathbb{E}_{x \sim \mathcal{X}} (\Delta_{\mathsf{adv}}(f_c,\trc,x)),
\end{align*}
where
\begin{align*}
  \Delta_{\mathsf{adv}}(f_c,\trc,x)=\min\limits_{r\in \R^{n_x}} \|r\|_2 \\
  \mathsf{s.t.} \quad \arg \max f_c(x+r) \neq \arg \max \trc(x).
\end{align*}
The value of $\rho_1$ is an important metric for the investigation of ML models \cite{mickisch_understanding_2020}. However, it only provides the mean distance of $x \sim \mathcal{X}$ to the decision boundary. Neither the severity of the misclassification on the application nor the likelihood of occurrence of the critical perturbation $r$ are considered.

A related property is the so called \emph{cross Lipschitz extreme value for network robustness} (CLEVER) score of a classifier introduced in \cite{weng2018evaluating}. For CLEVER, the maximum of the norm of the gradient in a ball around a test input value $x$ is estimated because it can be used to predict the distance of $x$ to the decision boundary. To arrive at the CLEVER score, the mean value over the maximal gradients in balls around $x \sim \mathcal{X}$ is computed.

In \cite{madry_towards_2019} a loss function $L$ is used which allows to quantify the effect of misclassification in the environment of the ML model under test. Using $L$, the authors define the \emph{adversarial risk}
\begin{align*}
  \rho_2(f_c,\mathcal{X},\trc) := \mathbb{E}_{x \sim \mathcal{X}} \left( \max \limits_{r \in \mathcal{S}} L(f_c(x+r),\trc(x)) \right),
\end{align*}
where $\mathcal{S}$ is a set of admissible perturbations. This definition works well when evaluating ML models in the adversarial setting to assess the mean of the maximal damages 
an adversary could potentially have on the deployed ML application in a specified environment. However, for general risk assessment this worst case definition does not apply. For that, we need a definition that also takes the probability of each perturbation into account.

To find such a definition, we turn to statistical decision theory and view the risk of deploying a ML model as the risk of a statistical decision making process.

\begin{defi} \emph{Risk of a statistical procedure \cite{berger_statistical_1985}
} \\ Let $\mathcal{X}$ be a probability space defined as above and let $\mathcal{A}$ be an action space. Furthermore, let $d:X  \rightarrow \mathcal{A}$ be a deterministic decision function. Then the risk of $d$ with respect to a loss $L$ in the setting of $\mathcal{X}$ is defined as
  \begin{align*}
    R(d)=\mathbb{E}_{x \sim \mathcal{X}} L(d(x))=\int \limits_X L(d(x)) \dd \mathbb{P}(x).
  \end{align*}
  For a randomized decision function $d^* : X \times \mathcal{D}_x  \rightarrow \mathcal{A}$ with the parametric probability space $\mathcal{D}_x=(N,\mathcal{G},\mathbb{P}_x)$ we have that
  \begin{align*}
    R(d^*)=\int \limits_X \int \limits_N L(d^*(x,\delta)) \dd \mathbb{P}_x(\delta) \dd \mathbb{P}(x)
  \end{align*}
  \label{def:stat_risk}
\end{defi}
This definition of risk for a deterministic decision function is well-suited for risk assessment of a ML model on unperturbed (test) data. On the other hand, the double integral formula is a good starting point for general ML risk assessment since it allows to cover both the original data distribution and possible perturbations. In the following we explain how this definition of risk can be applied to evaluate ML models.

\section{Risk Definition for Machine Learning Applications}

To utilize the risk definition from decision theory in ML, we translate all terms from Definition~\ref{def:stat_risk} to the ML domain. Our starting point is the randomized setting with the decision function $d^*$. First, we propose to decompose $d^*$ into a deterministic and a stochastic part, which represent the ML model and the input noise, respectively. Note that some ML models include randomization such as in some proposed adversarial defenses (\cite{xie_mitigating_2018} and \cite{meng_magnet_2017}). This additional randomization that is part of the ML model and that is not caused by input noise can be encompassed similarly by decomposing the ML model into a deterministic and a randomized part. Then for the evaluation of a randomized ML model a third integral is added.

After that decomposition, $\mathcal{X}$ and $\mathcal{D}_x$ immediately carry over to the ML setting. $\mathcal{X}$ represents the underlying data distribution and $\mathcal{D}_x$ represents natural and artificial noise. The interpretation of the loss and the decision function depend strongly on the specific use case. When the ML model is deployed to autonomously take actions, then the ML model is directly the decision function and the loss can simply rate the ML model's decisions. However, if the ML model is used for data analysis and only implies decisions within a more complex system we must either introduce a function that maps the ML model's output to a decision or incorporate the cost associated with worse decisions caused by faulty data analysis into the loss function. We propose to use the latter approach since this reduces the overall complexity of the evaluation.

Using the above considerations, we define the risk of deploying a classifier $f_c$ in an environment $\mathcal{X}$ with perturbations $\mathcal{D}_x$ by
\begin{align}
  \label{eq:our_loss}
  \rho(f_c,\mathcal{X},\mathcal{D}_x)=\int \limits_X \int \limits_N L(f_c(x+\delta))
  \dd \mathbb{P}_x( \delta) \dd \mathbb{P}(x),
\end{align}
where $L$ is a loss function that maps the classification to the loss of the resulting decision. Note that a possible explicit dependence of $L$ (an thus $\rho$) on $\trc$ and $x$ is omitted in \eqref{eq:our_loss}. Furthermore, in the adversarial setting, $\mathcal{D}_x$ can also depend on $f_c$.

Before explaining how \eqref{eq:our_loss} can be approximated, we give a few examples to roughly sketch the scope of our definition of risk.

The adversarial risk from \cite{madry_towards_2019} is encompassed by our framework. This can be seen by choosing $L$ as training loss, and $\mathcal{D}_x$ as space of adversarial perturbations computed as in \cite{madry_towards_2019} that occur for the given target image $x$ with probability one.

When an adversarial defense is proposed, the robustness evaluation is normally performed by checking the decrease in accuracy for different perturbation budgets. In our framework, this translates to a computation of $\rho$ with
\begin{align*}
  L(f_c( x+ \delta))=\mathbbm{1}_{\arg \max f_c(x+\delta)= \arg \max \trc(x)},
\end{align*}
for different noise distributions $\mathcal{D}_x$, where $\mathbbm{1}$ is the indicator function. Note that in this setting, to address different perturbation budgets, we can compute the risk multiple times for all different perturbation budgets.

We now describe the use case that is the main motivation for this work. When choosing a ML model as vision system for a self-driving car, it must be determined which model leads to the minimal risk when deployed. To assess the risk associated with deploying a ML model, the environment in which it is deployed is described using the natural distribution of input images $\mathcal{X}$ and the noise $\mathcal{N}$. As an example, the model might be deployed in an urban area (which is described by $\mathcal{X}$), in which fog and rain occur regularly and, moreover, there is a 0.1\% chance for an adversarial perturbation created with a transfer attack on one of the street signs (which is covered by an appropriate choice of $\mathcal{N}$).

Furthermore, a loss function that estimates the possible damage of a segmentation output is defined. A detailed description of such a loss function is beyond our scope. However, it is important to note that a simple measure such as the sum of misclassified pixels does not necessarily reflect the possible damage. A pedestrian not being detected on non-drivable area is less taxing than a pedestrian being missed on an area that is otherwise classified as drivable.

We emphasize the fact that our risk definition for ML applications via the double integral over the natural data distribution and the (possibly adversarial) noise allows a realistic description of the environment in which the ML model is deployed. On top of that, the loss function within the risk definition can be designed to weight each classification error based on its severity with respect to the given applications.

\section{The Key Risk Indicator Tensor}
\label{sec:kritensor}

We now turn to the computation of $\rho$ for given $L, \mathcal{X}$, and $\mathcal{D}_x$. For that, we propose to approximate the double integral \eqref{eq:our_loss} using a Monte Carlo simulation such that we have
\begin{align*}
  \rho \approx \hat \rho = \frac{1}{n_x n_{\delta}} \sum \limits_{i=1}^{n_x} \sum \limits_{j=1}^{n_\delta} L(f_c(x_i+\delta_j)),
\end{align*}
where $x_i$ and $\delta_j$ are samples from $\mathcal{X}$ and $\mathcal{D}_x$, respectively. This straightforward approach works well for fixed $L$, $\mathcal{X}$, and $\mathcal{D}_x$. However, when $\mathcal{X}$ or $\mathcal{D}_x$ are changed (e.g.\ if new scenarios are added), all computations have to be carried out again which is computationally taxing. Therefore, we propose a light-weight data structure from which $\hat \rho$ can be extracted that allows for more flexibility.

The basis for reusing inference results of a classifier $f_c$ for different deployment scenarios in which $\hat \rho$ is evaluated is the composition of the scenarios from sets $\{\mathcal{X}_i\}_{i=1}^{\nx}$ and $\{\mathcal{D}_{xi}\}_{i=1}^{\nd}$. Then a set of risk values $\{\hat \rho_i\}_{i=1}^{\nx \nd}$ can be computed and the final risk value $\hat \rho$ can be obtained as a convex combination of the elements $\hat \rho_i$ as
\begin{align}
  \label{eq:convexcomb}
  \hat \rho = \sum \limits_{i=1}^{\nx \nd} \alpha_i \hat \rho_i, \text{ with } \sum \limits_{i=1}^{\nx \nd}\alpha_i=1,
\end{align}
where $\alpha_i$ can be used to weight different scenario components from which the deployment scenarios are constructed. Note that all different $\hat \rho_i$ (and therefore $\mathcal{X}_i$ and $\mathcal{D}_{xi}$) need not be known at the same time. On the contrary, scenario components can be added later to further refine the description of the deployment scenario.

Another advantage of separating $\hat \rho$ into different components is a more detailed insight into failure modes of the ML model under test. When $\hat \rho$ is directly computed, we obtain no information on which parts of $\mathcal{X}$ or $\mathcal{D}_x$ have caused the risk to increase. However, this information is invaluable for uncovering weaknesses and improving the ML model. As an example, when the evaluation shows that a given segmentation model misses pedestrians in images that contain noise that mimics rain, this can initiate an analysis of whether images of that type are underrepresented in the training set or whether the given ML architecture can in general not deal with that type of noise. For that, we propose to interpret the different $\hat \rho_i$ as KRIs of an ML model.

A KRI describes the risk in one particular situation ($\mathcal{X}_i, \mathcal{D}_{xi}$) that may be part of the deployment scenario of the ML model. This can be obtained by modeling a part of the environment. Furthermore, in the adversarial setting, we can view a KRI as an indicator of the susceptibility of a given ML model to a particular adversarial attack. In this way, comparing different KRIs allows to analyze both the mode of action of different attacks as well as the failure modes of the ML model. 

When computing $\hat \rho_i$, our main objective is the reusability of the inference results of the ML model, since this is the computationally most expensive part. Therefore, we store the computation results in the risk tensor $R$ which is defined by
\begin{align*}
  R_{i,j,k,\ell}=L_i(f_c(x_j+\delta_{k,\ell})).
\end{align*}
$R$ is used to store the results for different loss functions $L_i$, samples of the natural distribution $x_j$, and different samples of a given noise distribution $\delta_{k,\ell}$. The different elements are joined along the different natural distributions and noise types to form the complete risk tensor $R$.

Note that a risk tensor $R^{\text{adv}}$ can be defined for the adversarial robustness use case. Since samples of adversarial noise are typically created for one specific input, we can reduce the dimension of the risk tensor and obtain
\begin{align*}
  R^{\text{adv}}_{i,j,k}=L_i(f_c(x_j+\delta_{k,j})).
\end{align*}
Here we have a one-to-one correspondence of the samples of the noise distribution to samples of the natural image distribution.

Once $R$ is computed, the different $\hat \rho_i$ can be obtained by filtering $R$ for distributions relevant for the specific situation which is encompassed by $\hat \rho_i$ and aggregating the different tensor elements. When all $\hat \rho_i$ have been computed, $\hat \rho$ can be obtained as in \eqref{eq:convexcomb}.

\section{Case Study}

\begin{figure}[tb]
  \centering
  \centering
  \includegraphics[width=1.0\linewidth]{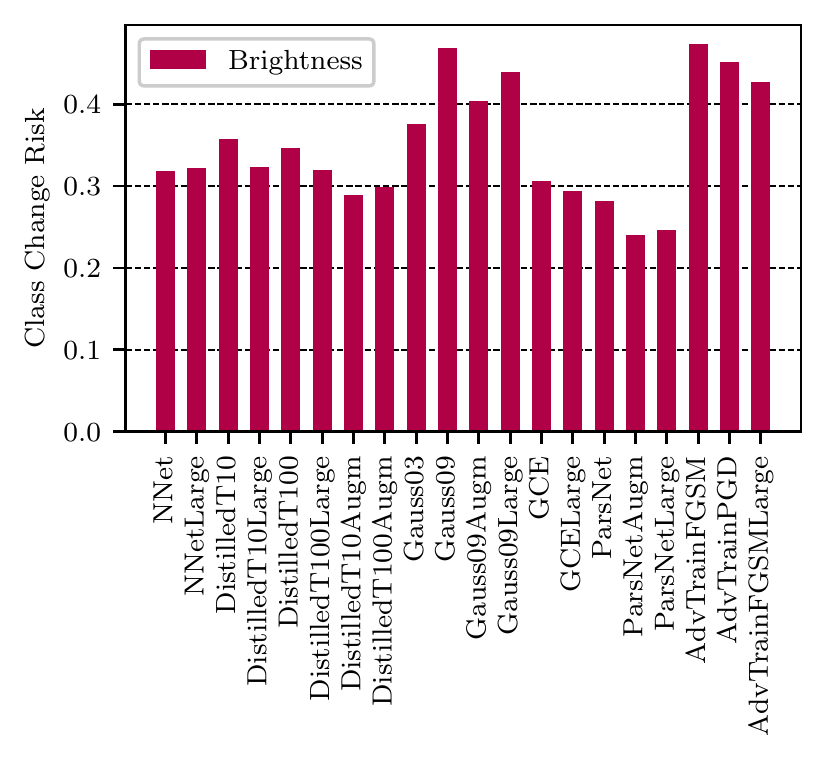}
  \caption{Comparison of class change risk for brightness perturbations}
  \label{fig:brightness}
\end{figure}
\begin{figure}[tb]
  \centering
  \includegraphics[width=1.0\linewidth]{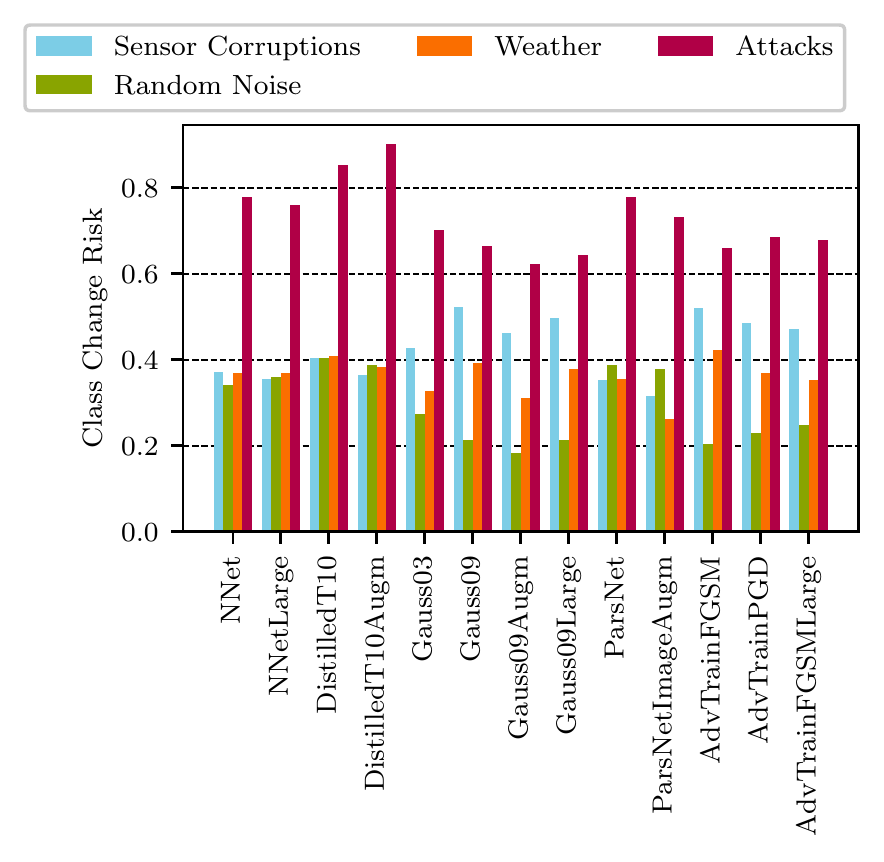}
  \caption{Assessment of the impact of image data augmentation for different noise types}%
  \label{fig:agr_data_aug}
\end{figure}

We demonstrate the feasibility and utility of our approach by computing the KRIs for a set of neural image classifiers. Note that the KRIs we use in this study are based on well-known and rather straightforward risk measures like the \emph{probability of class change} to keep our results well-aligned with state-of-the-art robustness investigations. In particular, we compare the KRIs of 20 residual neural networks (ResNets), trained on the CIFAR10 \cite{krizhevsky2009learning} dataset to investigate their respective robustness with respect to image corruptions and adversarial attacks.

We use different robustifying measures alongside changes in the ResNet depth to vary the ResNets under study. In particular, we vary the training data augmentations by adding both Gaussian noise and standard image augmentations implemented in Keras \cite{chollet2015keras}, adding a regularization proposed in \cite{cisse_parseval_2017}, changing the training loss function to the robustifying guided complement entropy loss \cite{chen_improving_2019}. Furthermore, we obtain a few ResNets from \emph{defensive distillation} as proposed in \cite{papernot_distillation_2016} performed at different distillation temperatures and by adversarial training (both \emph{ensemble adversarial training} \cite{tramer2018ensemble}, and \emph{projected gradient descent} \cite{madry_towards_2019} were tested). A description of the setup of each ResNet we study is provided in Table~\ref{tab:rnn_description}.

We evaluate and compare the ResNets' capability to cope with image perturbations induced by sensor corruptions, random noise, weather phenomena, and adversarial attacks. For each image perturbation type, we set up several distributions that represent each corruption scenario. For sensor corruptions, we consider random changes in brightness and contrast. On top of that, we add shadows and rotations of varying magnitude to the test images. Random noise is considered by adding distributions of Gaussian, uniform and salt-and-pepper noise. We incorporate weather phenomena by adding randomly generated layers of rain or fog to the test images. For the creation of adversarial perturbations, we use the adversarial robustness toolbox \cite{nicolae_adversarial_2019} to generate distributions that contain images created with the fast gradient sign method \cite{goodfellow_explaining_2015}, the CarliniL2 method \cite{carlini_towards_2017}, and the DeepFool attack \cite{moosavi-dezfooli_deepfool_2016}, respectively.

A key feature of our approach is the hierarchical aggregation of the computed loss values. In our study, the loss values are the probability of class changes, which can be aggregated with a mean value function over the different noise types. As an example, in Figure~\ref{fig:brightness}, we compare the class change risk of test images with brightness perturbations for the ResNets we study. These values are computed by
\begin{align*}
  \hat \rho_{\text{br}}= \frac{1}{n_\text{samples}}\sum\limits_{x_i \in X} \,\, \sum\limits_{\delta_{\text{br},\ell} \in \mathcal{D}_{\text{br}}} L_{cc} (x_i+\delta_{\text{br},\ell}),
\end{align*}
where $L_{cc}$ is the indicator for a class change, $X$ is the set of test images, and $\delta_{\text{br},\ell}$ is a sample from the distributions of brightness perturbations.

\begin{figure}[tb]
  \centering
  \includegraphics[width=1.0\linewidth]{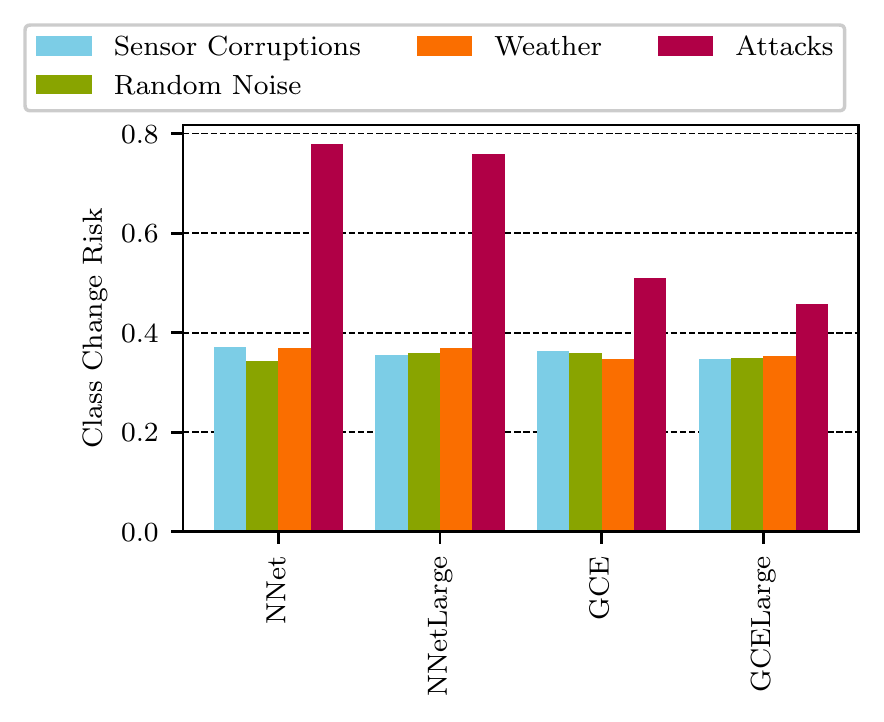}
  \caption{Assessment of the impact of the loss function for different noise types}%
  \label{fig:agr_gce}
\end{figure}

The individual values $\hat \rho_{\text{br}}$ for the different ResNets under study can be considered their brightness corruption KRIs. On the other hand, we can summarize the risk values for all sensor perturbations into a single sensor corruption KRI, by aggregating over all sensor corruption distributions. In this way, the sensor corruption KRIs are computed by
\begin{align*}
  \hat \rho_{\text{sc}}= \frac{1}{n_\text{samples}}\sum\limits_{x_i \in X} 
  \sum\limits_{D_i \in D_{\text{sc}}}
  \,\, \sum\limits_{\delta_{\text{i},\ell} \in {D}_{i}} L_{cc} (x_i+\delta_{i,\ell}),
\end{align*}
where $D_{\text{sc}}$ is the set of all considered sensor corruption distributions.
At this stage it is possible to weight the different corruptions in order to mimic their given occurrence probability. In Figure~\ref{fig:agr_data_aug} we use these higher level KRIs to understand the effect of data augmentation on the robustness of a ResNet. We can observe that by adding Gaussian noise and standard image augmentations or adversarial noise to the training images, we can increase the robustness of the ResNets with respect to random and adversarial noise by similar amounts. On the other hand, when we compare the standard cross entropy loss to the guided complement entropy loss as in Figure~\ref{fig:agr_gce}, we can observe that using the guided complement entropy loss, we can significantly increase the robustness with respect to adversarial noise. However, the vulnerability with respect to the other noise types stays approximately the same.

\begin{figure}[tb]
  \centering
  \includegraphics[width=1.0\linewidth]{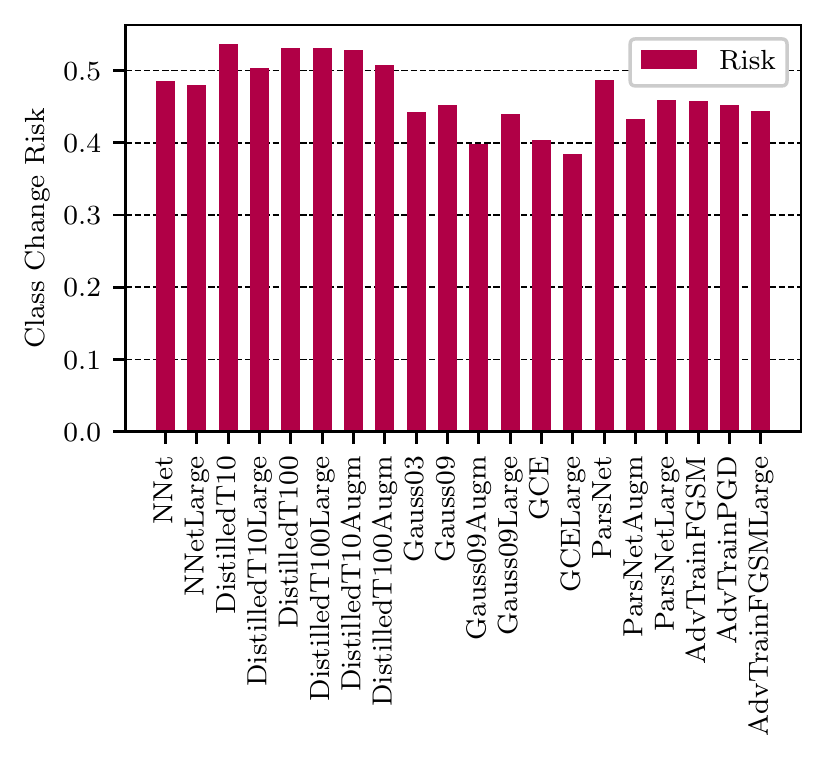}
  \caption{Final risk values}%
  \label{fig:final}
\end{figure}

Finally, we can merge all KRIs into the final risk value. In our study, we simply compute the mean value over all KRIs. However, a more involved strategy to study a specific use case can also be implemented. The final risk values are displayed in Figure~\ref{fig:final}. On the basis of these values, an informed choice of the ResNet associated with the minimal risk can be made.

\begin{table*}[tb]
  \centering
  \begin{tabular}{ll|cccc}
    \# & label              & \# layers & reg.           & augment.                 & defense       \\ \hline
    1  & NNet               & 72        & -              & -                        & -             \\
    2  & NNetLarge          & 114       & -              & -                        & -             \\
    3  & DistilledT10       & 72        & -              & -                        & distillation  \\
    4  & DistilledT10Large  & 114       & -              & -                        & distillation  \\
    5  & DistilledT100      & 72        & -              & -                        & distillation  \\
    6  & DistilledT100Large & 114       & -              & -                        & distillation  \\
    7  & DistilledT10Augm   & 72        & -              & std. image augmentations & distillation  \\
    8  & DistilledT100Augm  & 72        & -              & std. image augmentations & distillation  \\
    9  & Gauss03            & 72        & -              & Gaussian noise ($\sigma=0.03$)           & -             \\
    10 & Gauss09            & 72        & -              & Gaussian noise ($\sigma=0.09$)         & -             \\
    11 & Gauss09Augm        & 72        & -              & Gaussian noise ($\sigma=0.09$)        & -             \\
    12 & Gauss09Large       & 114       & -              & Gaussian noise ($\sigma=0.09$)       & -             \\
    13 & GCE                & 72        & -              & gce loss                 & -             \\
    14 & GCELarge           & 114       & -              & gce loss                 & -             \\
    15 & ParsNet            & 72        & parseval frame & -                        & -             \\
    16 & ParsNetAugm        & 72        & parseval frame & std. image augmentations & -             \\
    17 & ParsNetLarge       & 114       & parseval frame & -                        & -             \\
    18 & AdvTrainFGSM       & 72        & -              & -                        & adv. training \\
    19 & AdvTrainPGD        & 72        & -              & -                        & adv. training \\
    20 & AdvTrainFGSMLarge  & 114       & -              & -                        & adv. training \\
  \end{tabular}
  \caption{Description of ResNets used for KRI computation}
  \label{tab:rnn_description}
\end{table*}

\section{Conclusion and Outlook}

In this work, we have applied the risk definition from statistical decision theory to ML. On the basis of this definition we have developed a framework that allows to specify different deployment scenarios and penalties associated with failures of the ML model. This allows practitioners to evaluate the risk of deploying a given ML model in a standardized way. Furthermore, the setup of deployment scenarios gives regulatory authorities the chance design certificates for ML models in specified environments.

In our preliminary numerical case study we have provided another motivation for using KRIs to investigate ML model robustness, i.e.\ the investigation of the effect of different robustifying measures on perturbations of different types. As an example, while adding data augmentations increased the accuracy under random and adversarial perturbations, a change in loss function from cross entropy to guided complement entropy only increased robustness for adversarial perturbations.

The application of the risk definition and the proposal for its efficient tensor based evaluation provide the tools necessary for extensive analysis of ML models. It remains to create meaningful loss functions and data distributions for different applications in which such a detailed analysis is required.

\bibliography{mainKRI.bib}
\bibliographystyle{aaai}
\end{document}